\pdfoutput=1
\documentclass[11pt,a4paper]{article}
\usepackage{times,latexsym}
\usepackage{url}
\usepackage[T1]{fontenc}

\usepackage[]{lrec}
\date{} 
\usepackage{soul}

\usepackage{inconsolata}

\usepackage{graphicx}

\usepackage{fancyvrb}
\usepackage{fvextra}
\usepackage[shortlabels]{enumitem}

\usepackage{makecell}
\usepackage{multirow}
\usepackage{tabularx}
\usepackage{siunitx}
\usepackage{hyperref}

\title{Exploring In-Context Learning for Frame-Semantic Parsing}

\author{
  \textbf{Diego Garat},
  \textbf{Guillermo Moncecchi},
  \textbf{Dina Wonsever}
\\
  Facultad de Ingeniería,\\ Universidad de la República,\\ 11300 Montevideo, Uruguay
\\
  \small{
    \href{mailto:email@domain}{\{dgarat,gmonce,wonsever\}@fing.edu.uy}}
}

\begin{document}
\maketitle              

\begin{abstract}
{} \textbf{Abstact:}
Frame Semantic Parsing (FSP) entails identifying predicates and labeling their arguments according to Frame Semantics. This paper investigates the use of In-Context Learning (ICL) with Large Language Models (LLMs) to perform FSP without model fine-tuning. We propose a method that automatically generates task-specific prompts for the Frame Identification (FI) and Frame Semantic Role Labeling (FSRL) subtasks, relying solely on the FrameNet database. These prompts, constructed from frame definitions and annotated examples, are used to guide six different LLMs. Experiments are conducted on a subset of frames related to violent events. The method achieves competitive results, with $F_1$ scores of \qty{94.3}{\percent} for FI and \qty{77.4}{\percent} for FSRL. The findings suggest that ICL offers a practical and effective alternative to traditional fine-tuning for domain-specific FSP tasks.
\end{abstract}

\section{Introduction}
Frame Semantic Parsing (FSP) aims to detect and extract semantic frames as defined by Fillmore's Frame Semantics theory \cite{fillmore}, which posits that words evoke prototypical situations involving events, participants, locations, and times. The FrameNet Project,\footnote{\url{https://framenet.icsi.berkeley.edu/}} developed at the International Computer Science Institute in Berkeley, defines over \num{1220} semantic frames and more than \num{13600} lexical units for contemporary English \cite{baker-etal-1998-berkeley-framenet,baker-lorenzi-2020-exploring}. It also provides over \num{200000} manually annotated examples based on this theory.

The FSP task is typically divided into two subtasks: Frame Identification (FI) and Frame Semantic Role Labeling (FSRL). FI involves detecting the presence and type of frames, while FSRL focuses on identifying predicates and their arguments, labeling each argument with the appropriate semantic role. In this context, predicates are lexical units that evoke semantic frames, and their arguments are annotated as frame elements (FEs) \cite{gildea-jurafsky-2000-automatic}.

%% Since the early 2000s, FSRL has received considerable attention and has been featured in several shared tasks at major NLP conferences \cite{litkowski-2004-senseval, carreras-marquez-2005-introduction, baker-etal-2007-semeval}. A wide range of methods has been explored over the years, including rule-based systems, Support Vector Machines (SVMs), deep neural networks, machine reading comprehension techniques, and pre-trained language models \cite{roth-lapata-2015-context, bastianelli2020, Zheng_Wang_Chang_2023}.

Large Language Models (LLMs) have recently shown strong performance across diverse NLP tasks, often without modifying their architecture or parameters \cite{RADFORD19,BROWN2020,WEI24}. In in-context learning (ICL), a task description and input-output examples are provided at inference time to guide the model’s predictions. The number of examples can vary from a few (few-shot) to thousands (many-shot) \cite{BROWN2020,AGARWAL2024}.

This work investigates the effectiveness of applying ICL to the FSP task. Specifically, we examine whether LLMs can leverage their pre-trained knowledge to perform this task and how their performance is influenced by the number of provided examples. Our approach relies exclusively on the frame definitions and annotations from the FrameNet project, and we evaluate it using six models: Claude Haiku 3.5, Claude Sonnet 3.5, GPT-4o, GPT-4o Mini, Deepseek Chat, and Deepseek Reasoner.

\section{FrameNet}\label{section-framenet}
In the FrameNet project, each frame is described by a definition and its possible frame elements (FEs). These FEs are categorized into core and non-core elements: core FEs are essential to the meaning of the frame, while non-core FEs are generally circumstantial.

%% For instance, the Killing frame\footnote{\url{https://framenet.icsi.berkeley.edu/fnReports/data/frame/Killing.xml}} is defined as:
%% 
%% \begin{quote}
%%     A Killer or Cause causes the death of the Victim.
%% \end{quote}

%% The core FEs for this frame are \emph{Killer}, \emph{Cause}, \emph{Victim}, \emph{Instrument} and \emph{Means}, which are defined as follows:

%% \begin{itemize}
%%     \item \emph{Killer}: The person or sentient entity that causes the death of the Victim.
%%     \item \emph{Cause}: An inanimate entity or process that causes the death of the Victim.
%%     \item \emph{Victim}: The living entity that dies as a result of the killing.
%%     \item \emph{Instrument}: The device used by the Killer to bring about the death of the Victim.
%%     \item \emph{Means}: The method or action that the killing agent uses to provoke the death of the victim.
%% \end{itemize}

For instance, the Killing frame\footnote{\url{https://framenet.icsi.berkeley.edu/fnReports/data/frame/Killing.xml}} is defined as ``A Killer or Cause causes the death of the Victim''. The core FEs for this frame are \emph{Killer}, \emph{Cause}, \emph{Victim}, \emph{Instrument} and \emph{Means}, while non-core FEs include entities such as the beneficiary, time, and place. In addition, there is a special attribute called \emph{Target}, which signals the presence of the frame itself. The Target is represented in each sentence by a lexical unit that \emph{evokes} the frame. Figure~\ref{fig:killing-ex} shows an example for the  Killing frame.

\begin{figure}[t]
    \begin{center}
    \parbox{.45\textwidth}{
    \textcolor{blue}{They\textsuperscript{\emph{Killer}}} had \textbf{ killed\textsuperscript{\emph{Target}}} or captured \textcolor{red}{about a quarter of the enemy’s known leaders\textsuperscript{\emph{Victim}}}.
    }
    \end{center}
    \caption{Annotation example of the Killing frame, extracted from FrameNet.}  
    \label{fig:killing-ex}
\end{figure}

All examples in FrameNet database are classified by lexical unit, with each example consisting of a single sentence. Although a sentence may contain multiple instances from the same frame or different frames, in these examples only one frame instance is marked per sentence.
\section{In-Context FSRL}\label{section-in-context-fsrl}
%%Current LLMs often struggle to extract complex semantic relations without explicit guidance. To address this limitation, and with the purpose of extracting FrameNet relations, we propose using ICL to inject FrameNet knowledge into LLMs, avoiding the need for additional training. 

Current LLMs often struggle to extract complex semantic relations without explicit guidance. ICL is a paradigm that enables LLMs to learn how to solve a task by: (a) defining the task in natural language, and (b) providing input-output examples in the form of demonstrations. The primary advantage of ICL are the absence of the need for fine-tuning, which can reduce the size of the required training datasets \cite{DONG2024,BROWN2020}. %%However, this often comes at the cost of achieving lower performance compared to fine-tuned models \cite{DONG2024, BROWN2020}.

%%Our primary goal is to apply ICL to solve the Frame-Semantic Role Labeling (FSRL) task by leveraging the knowledge available in the FrameNet Project. To achieve this, our solution automatically generates the \emph{Prompt} for the LLM for a given task. This prompt includes the task description and demonstration examples derived solely from the Frame definitions and Lexical Unit (LU) examples. The prompt, combined with new input texts, is then fed into the LLM to produce the corresponding task outputs. Figure~\ref{incontext:fsrl_solver} illustrates the overall solution architecture.

\begin{figure}
\begin{center}
\includegraphics[width=.45\textwidth]{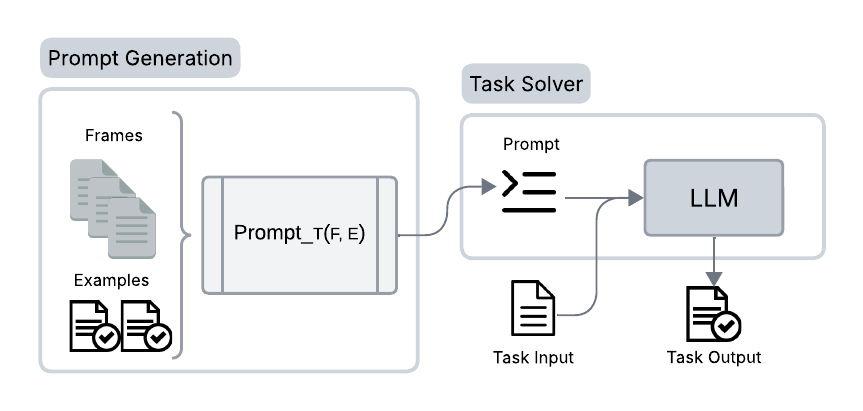}
\end{center}
\caption{In-context FSRL: FrameNet frames and lexical units, the annotated examples, are combined to construct a task-specific prompt. This prompt, along with an input text, is provided to the LLM to generate the corresponding task output.}
\label{incontext:fsrl_solver}
\end{figure}

Our primary goal is to apply ICL to solve the FSP task by leveraging the knowledge available in the FrameNet Project. Our solution automatically generates the LLM prompts including the task description and demonstration examples derived solely from the Frame definitions and Lexical Unit (LU) examples. A prompt, combined with new input texts, is then fed into the LLM to produce the corresponding task outputs. Figure~\ref{incontext:fsrl_solver} illustrates the overall architecture of the solution.

The primary limitation of this approach lies in the input length restrictions inherent to LLMs, which caps the number of frames and examples that can be included in a single prompt. Furthermore, the number of usable examples is constrained by the availability of annotated lexical units (LUs) for the different frames in the FrameNet project. In order to limit the prompt size, we propose to reduce the prompt generation to a subset of all frames available in FrameNet.

%% In order to limit the prompt size, we propose to reduce the prompt generation to a subset of all frames available in FrameNet. The drawback of this approach is that, for detecting any frame, the LLM must be called several times, each time with a new prompt that covers a new subset of frames until all frames are covered.

%In other words, given the set $\mathcal{F}$ of all available frames, and a set $E$ of annotated examples $x_i$ for task $T_i$, our prompt generator for each task can be seen as a function:     
%   $$Prompt_{T_i}: 2^\mathcal{F} \times 2^E \mapsto String $$

%%The number of examples used as shots affects only the size of the prompt, not the number of LLM calls. 
Each test item is processed individually: a test dataset containing $n$ elements results in exactly $n$ calls to the LLM, regardless of how many examples (shots) or frames are injected into the prompt. While batching multiple test items into a single prompt could reduce the number of calls, this strategy is intentionally avoided to prevent potential interactions between examples that could bias the model's outputs.

%% Section~\ref{sub-section-framenet-extraction} presents a range of tasks that explore alternative strategies for information extraction. One strategy attempts to extract complete relations using a single prompt, whereas others decompose the task into sequential subtasks—first identifying frames, then extracting their corresponding arguments—to address the problem in a step-by-step manner.

%% Section~\ref{sub-section-framenet-extraction} presents a range of tasks that explore alternative strategies for information extraction. Section~\ref{section-framework} details a solution that dynamically generates prompts based on a flexible template. This template incorporates definitions and examples from any specified list of input frames and is adapted according to the specific subtask—such as frame detection or argument extraction—being performed.

%% Section~\ref{sub-section-framenet-extraction} presents a range of tasks that explore alternative strategies for FSRL. Section~\ref{section-framework} introduces a solution that generates prompts from a flexible template, incorporating frame definitions and examples. This template is adapted to the specific requirements of each subtask.

\subsection{Frame-Semantic Parsing Tasks}\label{sub-section-framenet-extraction}
%% For frame parsing tasks, we implement a solution that dynamically generates prompts based on a template, incorporating definitions and examples from any given list of input frames. The template is adapted according to the specific sub-task --such as frame detection or argument extraction—that we aim to address.

%% A key limitation of this approach lies in the restricted input size of LLMs. The number of frame definitions and examples that can be included in a prompt is constrained by the model’s maximum input length, which limits the number of frames that can be processed in a single query. Consequently, this prevents the complete FSRL task from being performed for all frames in FrameNet within a single LLM execution.

We explore two alternative strategies. The first addresses the task in a single step: given a text, the LLM is expected to identify and label all detected frames for a predefined set of frames.  The second strategy divides the task into two stages: (a) \emph{Frame \& Target Identification}, where the LLM detects the frames evoked in the input text along with their lexical triggers\footnote{ This first stage combines the \emph{Target Detection} and \emph{Frame Identification} subtasks commonly found in prior work \cite{roth-lapata-2015-context,bastianelli2020}.}; and (b) \emph{Argument Identification}, where the identified frames and targets are passed back to the LLM to guide the extraction of their corresponding arguments.

Therefore, from the LLM perspective, we have three different types of tasks:

\begin{enumerate}
    \item \emph{Frame-Semantic Parsing (FSP)}: Detects all frames and their corresponding arguments in a text in a single call for a specific frame. The input is a text, and the output is a list of detected frames with their arguments.

%%       \begin{Verbatim}[frame=single,  samepage=true, fontsize=\small, breaklines=true, commandchars=\\\{\}]
%% \textbf{Text}: They had killed or captured about a quarter of the enemy's known leaders. 
%% \textbf{Output}: [\{
%%    "target": "killed", 
%%    "killer": "They", 
%%    "victim": "about a quarter of the enemy's known leaders."
%% \}]
%%    \end{Verbatim}

    \item \emph{Frame \& Target Identification (FI)}: Identifies Targets associated with different frames in a text. The input is a text, and the expected output is a list of frame-target pairs. %% This task is a combination of  Target Detection and Frame Identification sub-tasks used in other works \cite{roth-lapata-2015-context, bastianelli2020}.
    
%%    \begin{Verbatim}[frame=single,  samepage=true, fontsize=\small, breaklines=true, commandchars=\\\{\}]
%%\textbf{Text}: They had killed or captured about a quarter of the enemy's known leaders. 
%%\textbf{Output}: [("Killing", "killed")]
%%    \end{Verbatim}
    
%%    This task proved to be easier to solve, as this approach demonstrated strong performance, achieving good results across multiple frames simultaneously without requiring a large number of examples.

    \item \emph{Argument Identification (FSRL)}: Identifies all arguments from a given text based on a list of frames and targets. The input consists of a text and a list of frame-target pairs, and the output is the input list completed with the arguments. 
    
%%    \begin{Verbatim}[frame=single,  samepage=true, fontsize=\small, breaklines=true, commandchars=\\\{\}]
%%\textbf{Text}: They had killed or captured about a quarter of the enemy's known leaders. 
%%\textbf{Targets}: [\{ "frame": "Killing", "target: "killed" \}]
%%\textbf{Output}: [\{
%%    "target": "killed", 
%%    "killer": "They", 
%%    "victim": "about a quarter of the enemy's known leaders."
%%\}]
%%    \end{Verbatim}    
    
\end{enumerate}

In the FSP task, since the frames present in the input are unknown, the prompt must include examples for all selected frames, regardless of their actual presence in the text. This limits the number of examples per frame and restricts the ability to query multiple frames simultaneously. In contrast, the FSRL task benefits from prior FI, allowing the prompt to be tailored specifically to the detected frames, with definitions and examples adapted accordingly.

%% All tasks were evaluated using an increasing number of examples to assess the impact of shot count on LLM performance. In particular, for the third task, two assessments were conducted: one using the list of frame targets from the ground truth, ensuring that the input contains no errors apart from any potential inaccuracies made by FrameNet annotators; and another using the output from Task 2, generated by the same LLM, thereby propagating any mistakes made in the previous step. The latter case was evaluated using only 150-shot examples.

\subsection{Prompt Generation}\label{section-framework}
Prompts are generated automatically for each task using frame definitions from FrameNet and by selecting test examples based on the required number of shots. These components are formatted using a Jinja2 template,\footnote{Jinja2 is a template rendering engine for Python. See: \url{https://palletsprojects.com/p/jinja/}} which contains the task definition and the rules for rendering the various sections of the prompt.

We define three distinct prompt templates, corresponding to the tasks described in Section~\ref{sub-section-framenet-extraction}. Each template consists of four main sections: Goal, Events, Guidelines, and Examples. %%Appendix~\ref{app-prompt-template} provides a template along with a rendering example.

The Goal and Guidelines sections contain only minimal references to specific frames. The Goal section is the simplest, stating the general objective of the task. The Guidelines section provides detailed instructions on how to perform the task while constraining the output format to reduce errors and facilitate downstream processing. %% For example, it may enforce output length restrictions, prohibit additional explanations to minimize verbosity, and impose other constraints to ensure consistency.

In contrast, the Events and Examples sections are frame-specific and built from frame definitions and annotated examples. The Events section includes each frame's definition, an illustrative sentence, and a list of all associated FEs, each with its corresponding definition. The Examples section contains input--output pairs (i.e., the example shots) drawn from the annotated dataset. While the number of examples may vary, it is ultimately constrained by the model's input size and the availability of examples in FrameNet.

\section{Experimental Settings}\label{section:experimental_settings}
The objectives of these experiments are twofold: first, to assess the performance and feasibility of our approach, and second, to evaluate the impact of the number of shots in the prompt on model performance in the FSP task. Table~\ref{experimental_settings:experiments} summarizes the setup for each experiment.

\begin{table*}
\begin{center}
    \begin{tabular}{|c|l|c|c|c|}
        \hline
        \textbf{Exp.} & \textbf{Task} & \textbf{\# Examples} & \textbf{Models} & \textbf{Frames} \\
        \hline
        1  & Frame-Semantic Parsing & 0--400  & \makecell{Claude,\\GPT} & Killing \\ \hline
        2  & Frame \& Trigger Ident. &\multirow{2}{*}{{0--150}} & \multirow{3}{*}{\makecell{Claude,\\Deepseek,\\GPT}} & \multirow{3}{*}{\makecell{Abusing,Killing,\\Rape, Robbery,
\\Shoot Projectiles, Violence}}\\\cline{1-2}
  %      \hline
        3a  & Argument Ident. (gold) &   &   &  \\ \cline{1-3}
   %     \hline
        3b  & Argument Ident. (detected) & 150 & &  \\ \cline{1-4}
        \hline
    \end{tabular}
    \caption{Overview of the experiments. Claude, GPT and Deepseek refers to both models tested for each (Haiku 3.7 and Sonnet 3.7, GPT-4o and GPT-4o mini, and Chat and Reasoner, respectively).}
    \label{experimental_settings:experiments}
\end{center}
\end{table*}

While our framework is designed to be applicable to any set of frames, the experiments focus on a subset related to violent events. Specifically, we consider the frames Abusing, Killing, Rape, Robbery, Shoot Projectiles, and Violence\footnote{A detailed list of frames can be found at: \url{https://framenet.icsi.berkeley.edu/frameIndex}}. This selection is solely due to prompt size limitations and is otherwise arbitrary. Only core attributes are targeted in our experiments.\footnote{For instance, in the Killing frame, we consider Target, Killer, Cause, Victim, Instrument, and Means.}

For the FSRL task, evaluation is conducted through two sets of experiments. Both use the same evaluation texts but differ in how the input Frame-Target pairs are generated. In the first, these pairs are derived from the annotated dataset, while in the second, they are the output of the FI task. The latter serves as an end-to-end evaluation, assessing the entire pipeline as an FSP task resolver.

\subsection{Evaluation Metrics}
Following previous work, we adopt Precision ($P$), Recall ($R$), and $F_1$ scores using a \emph{micro} evaluation approach, where metrics are computed for individual arguments (e.g., Target, Agent, Victim).\footnote{Unlike a \emph{macro} evaluation, which considers an instance correct only if all its attributes are correctly identified.}

The evaluation is \emph{strict}, meaning entity boundaries must be precisely detected. Dataset inconsistencies, such as variations in the inclusion of frame elements, are not accounted for. While methods like argument head matching could mitigate such inconsistencies, they are not applied in these experiments.

\subsection{Large Language Models}
For the first task, we evaluate four different LLMs --Claude Haiku 3.5, Claude Sonnet 3.5, GPT-4o Mini, and GPT-4o--. For the rest, we expand the evaluation to include DeepSeek Chat, and DeepSeek Reasoner. 

All LLM parameters are set to their default values, except for the temperature, which is set to \num{0.01}. This low temperature is chosen to minimize the randomness of the model's output and make the outcome more predictable. %%  This temperature setting remains unchanged throughout all experiments.

\subsection{Datasets} \label{section:experimental:datasets}
Two datasets are constructed for all experiments: one for prompt generation (ICL examples) and one for evaluating LLM outputs. To ensure a balanced distribution of targets, samples are selected using stratified sampling based on lexical unit frequency. A total of \num{150} examples are used for ICL and \num{100} separate examples for evaluation. 

All experiments share the same data partitions, except for the FSP task: a simplified scenario is considered, focused on a single frame, Killing, but with a larger number of training examples (\num{400}). %% Here, \num{400} examples are used for training and \num{100} for evaluation.

\subsection{Prompt Examples}
To evaluate the effect of the number of ICL examples on model performance, we test each model with an increasing number of examples. This applies to all tasks except for the FSRL task in the end-to-end setup, where the number of examples is fixed at \num{150} for both the FI and FSRL prompts.

For each task, the same example subsets, and thus the same generated prompts, are used across all models. Examples are added incrementally in a fixed order: the first $n$ examples remain consistent across all prompts containing at least $n$ shots. However, not all shot configurations are evaluated for every model. Larger models are tested with fewer configurations than smaller ones. The number of examples for smaller models is \numlist{0;5;10;25}, followed by increments of \num{25}, while larger models start from \num{0} and increase in steps of \num{50}.

%%In the zero-example condition, the Examples section is omitted from the prompt, leaving only the definition, a basic example, and an argument description to guide the model\footnote{See lines 39–52 in the template of Figure~\ref{fig:prompt-killing}. For instance, if no examples are included, lines 37–49 of Figure~\ref{fig:prompt-killing} would be removed.}. Additional examples are introduced gradually while preserving their order. That is, for any given model, the subset of the first $n$ examples remains identical across all prompts containing at least $n$ shots.  

%%For a given task, the same subsets of examples --and consequently, the same generated prompts-- are used across all tested models. However, not all shot conditions are evaluated for every model. Specifically, Larger models are tested with fewer shot conditions compared to smaller models.) are tested on a subset of the shot conditions applied to the smaller models (Claude 3 Haiku, GPT-3.5, GPT-4o Mini, and DeepSeek Chat). The number of examples increases in increments of 25 for smaller models and 50 for larger models, starting from 0, or from 0, 5, and 10, depending on the specific experimental setup.
\section{Results}\label{section-results}
As detailed in Section~\ref{section:experimental_settings}, a total of four experiments are conducted (see Table~\ref{experimental_settings:experiments}). The best $F_1$ results for these experiments are presented in Table~\ref{results:exps1-4:f1}. %% The following sections provide a detailed analysis of the results across four experiments.

\begin{table*}
\begin{center}
    \begin{tabular}{|l|c|c|c|c|c|}
        \hline
            & \textbf{Exp. 1}& \textbf{Exp. 2} & \textbf{Exp 3a} & \textbf{Exp 3b (FSRL)} & \textbf{Exp 3b (FI+FSRL)}  \\
        \hline
        Claude Haiku 3.5 & (275) 71.1 & (100) 90.9 & (75) 68.9 & 53.5 & 67.6 \\
        \hline
        Claude Sonnet 3.5 & (250) \textbf{82.9} & (150) 93.1 & (150) \underline{76.3} & \underline{68.6} & \underline{78.6}\\
        \hline
        DeepSeek Chat & - & (100) \textbf{94.3} & (150) 72.3 & 68.6  &  78.5 \\
        \hline
        DeepSeek Reasoner & - & (50) \underline{93.5} & (150) \textbf{77.4} & \textbf{74.9}   & \textbf{82.5} \\
        \hline
        GPT-4o Mini & (375) 68.7 & (125) 90.5 & (150) 67.1 & 53.2  & 66.8\\
        \hline
        GPT-4o & (400) \underline{80.2} & (100) 92.9 & (100) 73.9 & 66.9 & 78.0\\
        \hline
    \end{tabular}
    \caption{Micro $F_1$ across experiments. Best shots are shown for FSP (Experiment~1), FI  (Experiment~2) and FSRL with the true Frame-Target pairs  as input (Experiment~3a); FSRL with the output of FI (Experiment~3b) is evaluated for 150 shots. The highest scores are bold, while the second-highest scores are underlined.}
    \label{results:exps1-4:f1}
\end{center}
\end{table*}

Although a direct comparison with previous work is not feasible due to differing experimental setups, the results obtained in our experiments are nonetheless promising. Prior studies report $F_1$ scores for the FSRL task ranging between \qty{75.56}{\percent} and \qty{77.06}{\percent} on the FrameNet 1.7 dataset, using gold targets and frames \cite{roth-lapata-2015-context,bastianelli2020,Zheng_Wang_Chang_2023,Ai_Tu_2024}. In contrast, our best $F_1$ score reaches \qty{77.4}{\percent}, achieved by the DeepSeek Reasoner model with \num{150} shots.

In \cite{bastianelli2020}, the authors also report a \qty{76.8}{\percent} $F_1$ score for the Target Identification task and a \qty{90.1}{\percent} accuracy for Frame Identification, the latter assuming gold targets. In our joint FI task, we achieve a best $F_1$ score of \qty{94.3}{\percent}, with individual scores of \qty{95.3}{\percent} for Frame and \qty{94.3}{\percent} for Target Identification, using the DeepSeek Chat model with \num{100} shots.

\subsection{Experiment~1: FSP Task}
The first experiment involves detecting instances of the Killing Frame and their arguments. In this experiment, the number of shots varies between \numlist{0;400}, and four LLMs are evaluated: Claude Haiku 3.5, Claude Sonnet 3.5, GPT-4o mini, and GPT-4o. %% As in the other experiments, the number of shots is increased in increments of \num{25} for smaller models and \num{50} for larger models.

\begin{figure*}
\begin{center}
\includegraphics[width=.45\textwidth]{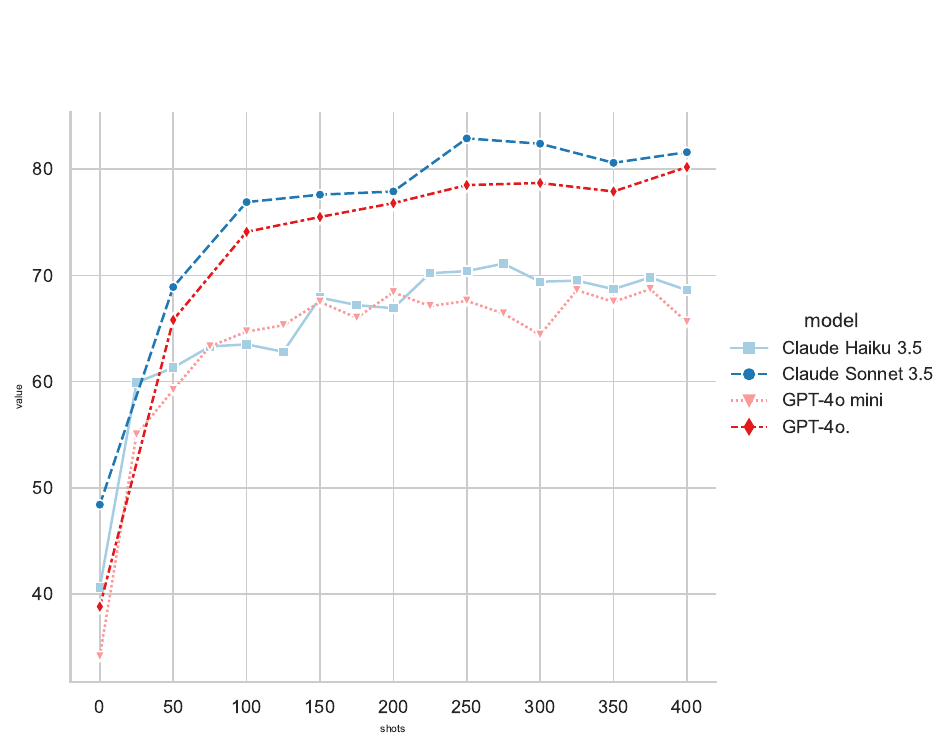}
\includegraphics[width=.45\textwidth]{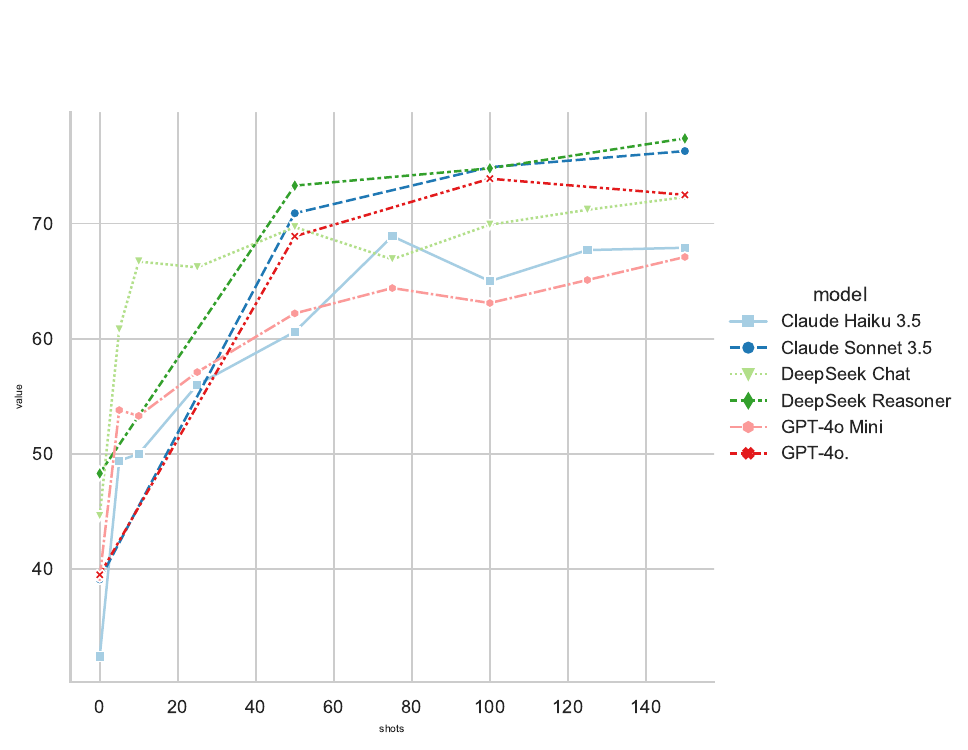}
\end{center}
\caption{Micro $F_1$ results by number of shots for Experiment~1 (left) and Experiment~3a (right).} \label{results:killing-arg_extraction:f1}
\end{figure*}

Figure~\ref{results:killing-arg_extraction:f1} (left) shows micro $F_1$ scores as a function of the number of shots, considering both Target and related attributes. Claude Sonnet achieves the highest performance at \num{250} shots, attaining an $F_1$ score of \qty{82.9}{\percent}, while GPT-4o reaches its peak of \qty{80.2}{\percent} at \num{400} shots. Sonnet consistently outperforms GPT-4o at all shot counts and surpasses GPT-4o’s peak performance beyond \num{250} shots. As expected, smaller models underperform compared to their larger counterparts. Claude Haiku achieves its highest $F_1$ score of \qty{71.1}{\percent} at \num{275} shots, while GPT-4o Mini peaks at \qty{68.7}{\percent} with \num{375} shots.

\begin{table*}
\begin{center}
    \begin{tabular}{|l|r|r|r|r|}
        \hline
             & \textbf{Haiku 3.5} & \textbf{Sonnet 3.5} & \textbf{GPT-4o Mini}  & \textbf{GPT-4o}\\
        \hline
        Target      & (200) 96.0 & (300, 350, 400) 96.6 & (200) 93.0 & (250, 300) \textbf{97.0} \\
        \hline
        Victim       & (225) 73.2 & (250, 300) \textbf{83.3} & (150) 72.6 & (400) 78.7 \\
        \hline
        Killer       & (300) 51.4 & (300) 68.7 & (375) 47.9 & (400) \textbf{71.4} \\
        \hline
        Cause        & (325) 45.8 & (250) \textbf{68.1} & (350) 43.5 & (150) 56.5 \\
        \hline
        Means        & (275) 22.2 & (300) \textbf{50.0} & 0.0 & (100) \textbf{50.0} \\
        \hline
        Instrument   & (275) 33.3 & (250, 400) \textbf{50.0} & 0.0 & (350,400) 22.2 \\
        \hline
    \end{tabular}
    \caption{$F_1$ scores for each attribute across different models for Experiment~1. Numbers in parentheses indicate the shot count at which the peak performance was achieved.}
    \label{results:killing:arguments:best}
\end{center}
\end{table*}

\begin{table*}
\begin{center}
    \begin{tabular}{|l|r|r|r|}
        \hline
             & \textbf{Name-Target} & \textbf{Name} & \textbf{Target}  \\
        \hline
        Claude Haiku 3.5 &(100)  90.9  & (100) 91.8 & (100) 90.9 \\
        \hline  
        Claude Sonnet 3.5 & (150) 93.1 & (150) 94.0  & (150) 93.1  \\
        \hline        
        DeepSeek Chat & (100) \textbf{94.3} & (100) \textbf{95.3}  & (10, 100) \textbf{94.3} \\
        \hline
        DeepSeek Reasoner & (50) \underline{93.5}   & (50) \underline{94.4} & (50) \underline{93.5} \\
        \hline
        GPT-4o Mini & (125) 90.5  & (125) 92.4  & (100) 91.0  \\
        \hline
        GPT-4o & (100) 92.9  & (100) 93.8 & (100) 92.9  \\
        \hline
    \end{tabular}
    \caption{$F_1$ for Experiment~2 by Pair and by its components Name and Target separately. Numbers in parentheses indicate the shot count at which the peak performance was achieved.}
    \label{results:framedetection:best}
\end{center}
\end{table*}

With the addition of examples, all four models show significant improvements in precision and recall compared to their zero-shot counterpart, with $F_1$ gains ranging from \num{30} to \num{41} points. This suggests that the knowledge embedded in the LLMs’ pretraining alone is insufficient to complete the task using only frame definitions.

Table~\ref{results:killing:arguments:best} summarizes the best results for each of the arguments. Target and Victim are the most accurately extracted arguments, achieving peak $F_1$ scores of \qty{97.0}{\percent} and \qty{83.3}{\percent}, respectively. For Target, larger models consistently score above \qty{95.0}{\percent} regardless of the number of positive shots. 

In a second performance tier are Killer and Cause. Killer reaches \qty{71.4}{\percent} with GPT-4o, while Cause peaks at \qty{68.1}{\percent} with Claude Sonnet. Smaller models score below \qty{51.4}{\percent} and \qty{45.8}{\percent}, respectively. 

Finally, performance on the remaining arguments is comparatively low. The best $F_1$ scores for these arguments reach only \qty{50.0}{\percent} for larger models and fall below \qty{22.2}{\percent} for smaller ones. Location is the only argument not annotated in any of the \num{100} evaluated examples.

%% In a second performance tier are Killer and Cause. GPT-4o and Claude Sonnet achieve $F_1$ scores of \qty{71.4}{\percent} and \qty{68.7}{\percent} for Killer, with smaller models trailing by approximately \num{20} points. For Cause, Claude Sonnet and GPT-4o reach \qty{68.1}{\percent} and \qty{56.5}{\percent}, while smaller models score below \qty{45.8}{\percent}.

%% Finally, performance on the Means and Instrument arguments is comparatively weak. The best $F_1$ scores for these arguments peak at only \qty{50.0}{\percent} for larger models, and fall below \qty{22.2}{\percent} for smaller ones. Location is the only argument not identified in any of the \num{100} evaluated examples.

\begin{figure*}[t]
\begin{center}
    \renewcommand{\arraystretch}{1.4}
    \begin{tabularx}{\textwidth}{l X X}
    a. &
    \textcolor{blue}{They\textsuperscript{\emph{Killer}}} tried to \textbf{assassinate\textsuperscript{\emph{Target}}} \textcolor{red}{her\textsuperscript{\emph{Victim}}}; and killed off two of her closest political friends, Airey Neave and Ian Gow. 
    & 
    \textcolor{blue}{They\textsuperscript{\emph{Killer}}} tried to \textbf{assassinate\textsuperscript{\emph{Target}}} \textcolor{red}{her\textsuperscript{\emph{Victim}}}; and \textbf{killed\textsuperscript{\emph{Target}}} off  \textcolor{red}{two of her closest political friends, Airey Neave and Ian Gow\textsuperscript{\emph{Victim}}}. 
    \\ \hline
    b. &
    After a while \textcolor{blue}{they\textsuperscript{\emph{Killer}}} kidnap and \textbf{murder\textsuperscript{\emph{Target}}} \textcolor{red}{a young boy\textsuperscript{\emph{Victim}}} for kicks, bashing him over the head with a blunt instrument. 
    & 
    After a while \textcolor{blue}{they\textsuperscript{\emph{Killer}}} kidnap and \textbf{murder\textsuperscript{\emph{Target}}} \textcolor{red}{a young boy\textsuperscript{\emph{Victim}}} for kicks, \textcolor{magenta}{bashing him over the head with a blunt instrument\textsuperscript{\emph{Means}}}. 
    \\ \hline
    c. &
    A professional assassin \textcolor{blue}{who\textsuperscript{\emph{Killer}}} climbed the walls and \textbf{murdered\textsuperscript{\emph{Target}}} \textcolor{red}{the woman\textsuperscript{\emph{Victim}}} \ldots %%without anyone catching sight of him.
    & 
    \textcolor{blue}{A professional assassin\textsuperscript{\emph{Killer}}} who climbed the walls and \textbf{murdered\textsuperscript{\emph{Target}}} \textcolor{red}{the woman\textsuperscript{\emph{Victim}}} \ldots 
    
    \end{tabularx}    
 
    \caption{
    Annotation examples of the Killing frame extracted from FrameNet (left), and made by Claude Sonnet 3.5 with 250 shots (right). In (a) there are two instances of Killing, but only one is marked in FrameNet; in (b) the model correctly identifies Means, missing in the annotated text; (c) shows an LLM error, where the model misses the pronoun \textit{who}, although the detected killer is semantically correct.   
    }
    \label{fig:annotation_comparative}
\end{center}
\end{figure*}

\begin{figure}[t]
\begin{center}
    \renewcommand{\arraystretch}{1.4}
    \begin{tabularx}{0.5\textwidth}{X}
    a. Last year \textcolor{red}{the president\textsuperscript{\emph{Victim}}} was \textbf{killed\textsuperscript{\emph{Target}}}  \textcolor{blue}{by rebels whom the Americans did not try to stop\textsuperscript{\emph{Killer}}}. \\ \hline
    b. He says \textcolor{blue}{Tutilo\textsuperscript{\emph{Killer}}} \textbf{killed\textsuperscript{\emph{Target}}} \textcolor{red}{ the man\textsuperscript{\emph{Victim}}} he tricked into helping him to steal away your saint.
    \end{tabularx}
    \caption{Examples extracted from FrameNet with different boundaries for relative clauses.}  
    \label{fig:annotation_errors}
\end{center}
\end{figure}

%% Some of these values, particularly for arguments other than Killer and Victim, arise from discrepancies with the gold-standard annotations. Figure~\ref{fig:annotation_comparative} illustrates some of these inconsistencies. Figure~\ref{fig:annotation_errors} provides another example, highlighting inconsistencies in how relative clauses are annotated in the dataset. These cases highlight potential annotation errors or omissions in the gold standard, emphasizing the importance of thorough validation, an aspect that falls beyond the scope of the present study.

Some values, especially for arguments other than Killer and Victim, arise from discrepancies in the gold-standard annotations. Figures~\ref{fig:annotation_comparative} and~\ref{fig:annotation_errors} illustrate inconsistencies, such as in the annotation of relative clauses. These issues point to potential errors or omissions in the gold standard, underscoring the need for thorough validation, an aspect that falls beyond the scope of this study.

\subsection{Experiment~2: FI Task}  
The second experiment evaluates model performance on the FI task, limited to the set of Violent Frames. All six LLMs are included, with the number of shots varying from \numrange{0}{150}. %% Specifically, we use \numlist{0;5;10} shots initially, followed by increments of \num{25} for smaller models and \num{50} for larger ones, starting from \num{25} and \num{50} shots, respectively.

Table~\ref{results:framedetection:best} reports the best $F_1$ scores for full Frame-Target pairs, as well as for Name and Target identification separately. Notably, all models achieve over \qty{90}{\percent} on Frame-Target extraction, with DeepSeek models leading at \qty{94.3}{\percent} and \qty{93.5}{\percent}.

Consistent with earlier results, all models show improved $F_1$ scores with more examples. However, given the already high initial scores, gains are modest, ranging from \num{1.4} to \num{8.7} percentage points.

\subsection{Experiment~3a: FSRL Task}  
In this experiment, the task is Argument Identification based on the input text and a list of gold-standard frame-target pairs derived from annotated data.

%% In this experiment, the model is tasked with identifying arguments based on the input text and a list of gold-standard targets and frames derived from annotated data. As in previous experiments, the number of shots begins at \numlist{0;5;10} and increases incrementally by \num{25} for smaller models and by \num{50} for larger models, starting at \num{25} and \num{50}, respectively.

Figure~\ref{results:killing-arg_extraction:f1} (right) shows $F_1$ scores by shot count. As summarized in Table~\ref{results:exps1-4:f1}, larger models achieve higher micro $F_1$ scores, ranging from \qty{73.9}{\percent} to \qty{77.4}{\percent}, while smaller models peak between \qty{67.1}{\percent} and \qty{72.3}{\percent}. As in Experiment~1, larger models outperform their smaller counterparts by \numrange{5}{7} points, although the margin is slightly narrower here.

%% \begin{figure}
%% \begin{center}
%% \includegraphics[width=.45\textwidth]{figs/micro_arg_extraction_F1_args.eps}
%% \end{center}
%% \caption{$F_1$ by number of shots for  Argument Detection with gold Frame-Target pairs (Exp. 3a).} \label{results:arg_extraction:f1}
%% \end{figure}

\subsection{Experiment~3b: FI + FSRL Tasks}  
Building on Experiment~3a, this experiment also addresses the FSRL task, but uses Frame-Target pairs predicted in Experiment~2, thereby evaluating full FSP via ICL. All six LLMs are included, each evaluated with \num{150} shots for both FI and FSRL tasks. No cross-model evaluation is conducted; each model's output serves as input to its own argument extraction.

%% Figure~\ref{results:end-to-end:f1} presents $F_1$ scores for 150-shot configurations in Experiments~2, 3a, and 3b. As expected, all models perform better in Experiment~3a, which uses gold-standard Frame-Target pairs, compared to the end-to-end setup in Experiment~3b. 

As expected, all models perform better in Experiment~3a, which uses gold-standard Frame-Target pairs, compared to the end-to-end setup in Experiment~3b. As shown in Table~\ref{results:exps1-4:f1}, the DeepSeek models demonstrate notable consistency, with only minor drops in $F_1$, \num{2.5} points for Reasoner and \num{3.7} points for Chat, while other models exhibit larger declines.

%%In contrast, other models exhibit larger declines, ranging from \num{5.6} points (GPT-4o) to \num{14.4} points (Claude Haiku).

%% \begin{figure}
%% \begin{center}
%% \includegraphics[width=.45\textwidth]{figs/micro_end-to-end_F1_args.eps}
%% \end{center}
%% \caption{$F_1$ with 150 shots for Frame and Target Detection (Exp. 2), and Argument Detection with gold (Exp. 3a) and detected Frame-Target pairs (Exp. 3b).} \label{results:end-to-end:f1}
%% \end{figure}

Beyond the FSRL task, the chained FI and FSRL setup also addresses the broader FSP objective of Experiment~1 within an ICL framework. Table~\ref{results:exps1-4:f1} shows $F_1$ scores reflecting both targets and arguments across all six models. Larger models achieve scores between \qty{78.0}{\percent} and \qty{82.5}{\percent}, while smaller ones range from \qty{66.8}{\percent} to \qty{78.5}{\percent}.

Since the set of frames and evaluation data differ from Experiment~1, direct comparison of results is not possible. However, even when evaluating on multiple frames instead of only the Killing frame, all models perform competitively with substantially fewer shots. %%—particularly GPT-4o, which uses three times fewer examples—showing only a \num{1.9} to \num{4.3} point drop in $F_1$ score.

\subsection{Ablation}
To assess the impact of FrameNet-derived information on model performance across tasks, we consider two additional prompting setups. In the first, prompts include only task descriptions and general instructions, without any FrameNet-specific content. In the second, prompts are extended to include the definition of the relevant frame, but omit any details or examples of its associated arguments. 

Table~\ref{results:ablation:f1} reports the results for these configurations, alongside those obtained using both frame and argument definitions: zero-shot prompting, and best-shot prompting with in-context examples.

%% \begin{figure*}
%% \begin{center}
%% \includegraphics[width=.30\textwidth]{figs/micro_ablation_task1_F1.eps}
%% \includegraphics[width=.33\textwidth]{figs/micro_ablation_task2_F1.eps}
%% \includegraphics[width=.33\textwidth]{figs/micro_ablation_task3_F1.eps}
%% \end{center}
%% \caption{$F_1$ of results for Frame-Semantic Parsing (upper), Frame \& Target Identification (middle) and Argument Identification (lower).} \label{results:ablation:f1}
%% \end{figure*}
%% \end{comment}

\begin{table*}[ht]
\centering
\begin{tabular}{llrrrr}
\hline
Task & Model & \makecell{a. Zero-shot\\No Frame Info} & \makecell{b. Zero-shot\\Frame Def.} & \makecell{c. Zero-shot\\ All Frame Info} & \makecell{d. Best-shot\\All Frame Info} \\

\hline
\multirow{4}{*}{\makecell{Exp.~1\\FSP}}
& Claude Haiku    & 0.0 & 0.0 & \underline{40.6}  & (275) \textbf{71.1} \\
& Claude Sonnet   & 0.0 & 0.0  & \underline{48.4} & (250) \textbf{82.9} \\
& GPT-4o Mini    & 0.0 & 0.0  & \underline{34.1}  & (375) \textbf{68.7} \\
& GPT-4o         & 0.0 & 0.0 & \underline{38.8} & (400) \textbf{80.2} \\
\hline
\multirow{6}{*}{\makecell{Exp.~2\\FI}}
& Claude Haiku      & 69.9 & 74.5 & \underline{83.7} & (100) \textbf{90.9} \\
& Claude Sonnet   & 82.2 & 83.1  & \underline{90.0} & (150) \textbf{93.1} \\
& DeepSeek Chat      & 90.4 & \underline{91.7} & 91.1 & (100) \textbf{94.3} \\
& DeepSeek Reasoner & \underline{92.6} & 92.2 & 92.1 & (50) \textbf{93.5} \\
& GPT-4o Mini       & 67.3 & 63.5 & \underline{81.8} & (125) \textbf{90.5} \\
& GPT-4o           & 74.1 & 71.3 & \underline{90.7} & (100) \textbf{92.9} \\
\hline
\multirow{6}{*}{\makecell{Exp.~3a\\FSRL}}
& Claude Haiku      & 22.2 & 33.0 & \underline{32.4} & (75) \textbf{68.9} \\
& Claude Sonnet     & 35.0 & \underline{39.1} & \underline{39.1} & (150) \textbf{76.3} \\
& DeepSeek Chat     & 28.0 & 42.4 & \underline{44.6} & (150) \textbf{72.3} \\
& DeepSeek Reasoner & 41.3 & \underline{50.9} & 48.3 & (150) \textbf{77.4} \\
& GPT-4o Mini    & 3.9 & 36.4 & \underline{39.3} & (150) \textbf{67.1} \\
& GPT-4o           & 25.6 & 32.1 & \underline{39.5} & (100) \textbf{73.9} \\
\hline
\end{tabular}
\caption{$F_1$ scores by model and prompting strategy across ablation for Experiments 1, 2 and 3a. 
(a) Zero-shot without frame information; 
(b) zero-shot with frame definitions only; 
(c) zero-shot with both frame and argument definitions; 
(d) best-shot with full frame and argument information, with the number of in-context examples shown in parentheses. 
For each task and model, the best score is in bold, and the second-best is underlined.}\label{results:ablation:f1}
%% \caption{$F1$ scores by model and prompting strategy across ablation experiments: (a) zero-shot with no frame information; (b) zero-shot with only the frame definition; (c) zero-shot with both frame and argument definitions; (d) best-shot (in parenthesis) with frame and argument definitions. Best result per model and task is in bold; second-best is underlined.}\label{results:ablation:f1}

\end{table*}

Overall, the experiments show that including frame definitions generally improves performance, especially for the FSP task. For simpler tasks, like FI task, DeepSeek models match or exceed zero-shot performance even with limited frame information. Notably, in the End-to-End task, DeepSeek Reasoner outperforms its zero-shot baseline when provided with only task and frame definitions. These results suggest that the benefit of frame information depends on task complexity and model capabilities.
\section{Related Work}
%%Over the past two decades, statistical methods for FSP have evolved significantly, spanning from SVM classifiers to advanced models such as LSTMs and pre-trained models such as BERT \cite{gildea-jurafsky-2000-automatic,baker-etal-2007-semeval,DAS2014,bastianelli2020,Zheng_Wang_Chang_2023, Ai_Tu_2024}. %%  A key resource in this field has been the FrameNet Project \cite{baker-etal-1998-berkeley-framenet}, which provides an extensive description of semantic frames, their attributes, relations, and annotated examples.
Since the early 2000s, FSRL has received considerable attention and has been featured in several shared tasks at major NLP conferences. A wide range of methods has been explored over the years, including rule-based systems, Support Vector Machines, machine reading comprehension techniques, and pre-trained language models \cite{gildea-jurafsky-2000-automatic,litkowski-2004-senseval,carreras-marquez-2005-introduction,baker-etal-2007-semeval,DAS2014,roth-lapata-2015-context}.

An example of a two-step solution can be found in \cite{bastianelli2020}: first they identify the spans in the input's constituency tree, training a Graph Convolutional Network,  and then they assign them a role label by using two CRFs, one for identifying arguments and the other to label them. They achieved a \qty{75.56}{\percent} of $F_1$ score in FrameNet 1.5 dataset with gold targets and frames.  

AGED \cite{Zheng_Wang_Chang_2023} is a query-based framework that exploits  definitions for frames in FrameNet: texts and frames definitions are concatenated and used as input examples to train a language model;  FE definitions are used as slots that should be filled by the extracted arguments. This framework gets a \qty{76.91}{\percent} of $F_1$ score in FrameNet 1.7 dataset.  In \cite{Ai_Tu_2024}, they follow the AGED model but, for each frame,  they build a Conditional Random Field to model the relation between frame arguments; \qty{77.06}{\percent}  of $F_1$ score in FrameNet 1.7 dataset.

%Although In-Context Learning (ICL), and particularly Many-Shot ICL, has been explored in several NLP tasks~\cite{DONG2024, AGARWAL2024}, such as summarization, sentiment analysis, and translation, to our knowledge, there are no existing experiments that apply this paradigm to FSP.

\section{Conclusions}
This paper introduces a framework for solving the FSP task using LLMs, leveraging the semantic information from the FrameNet Project via ICL. Although ICL has been explored in several NLP tasks~\cite{DONG2024,AGARWAL2024}, to our knowledge, there are no existing experiments that apply this paradigm to FSP.

%%To our knowledge, this is the first application of ICL to FSP.

Our framework automatically generates prompts for specific sub-tasks and frame sets, querying LLMs to detect frames, targets, and their attributes. Prompt generation relies exclusively on definitions and examples from the FrameNet database. Due to input length constraints, we evaluate the approach on a subset of FrameNet frames using six different LLMs on annotated texts.
%%Due to input length constraints, we evaluated our approach on a subset of FrameNet frames: \emph{Abusing, Killing, Rape, Robbery, Shoot Projectiles}, and \emph{Violence}, using six different LLMs on FrameNet-annotated texts.

All models showed improvements with task-specific examples, emphasizing that the models' intrinsic knowledge alone is insufficient to solve these tasks using only frame definitions and a single example. The degree of improvement varied based on both model type and task complexity.

Our experiments show that ICL can achieve competitive $F_1$ scores of \qty{94.3}{\percent} for FI and \qty{77.4}{\percent} for FSRL on a limited set of frames. This suggests that ICL is a viable alternative to training new models from scratch, particularly in domain-specific scenarios focusing on a small set of relevant frames.

While the results are promising, further fine-grained evaluation is needed to address annotation inconsistencies, including relative clause boundaries, missing attributes, and the single-instance-per-example limitation.

\bibliography{biblio}
\bibliographystyle{lrec}

%%\appendix

%%\input{sections/a1-prompt-example}

\end{document}